%% file: main.tex
\documentclass{article}

\input{settings}
\input{math_abbrev}

\begin{document}

\title{Equivalence of approximation by networks of single- and multi-spike neurons}

\author{
  Dominik Dold, Philipp Petersen\\[2pt]
  Faculty of Mathematics and Research Network Data Science @ Uni Vienna\\
  University of Vienna\\
  Kolingasse 14-16, 1090 Vienna, Austria \\
  \texttt{dominik.dold@univie.ac.at} \\
}

\maketitle
\begin{abstract}
\input{content/abstract}
\end{abstract}

\section{Introduction}

\input{content/introduction}

\section{Background}\label{section:background}

\input{content/background}

\section{Results}

\input{content/results}

\section{Limitations}

\input{content/limitations}

\section*{Acknowledgments}

\input{content/acknowledgments}

\printbibliography
\addcontentsline{toc}{section}{References}

\end{document}

%% file: settings.tex
\usepackage{microtype}

\usepackage{arxiv}

\usepackage[font=small,labelfont={bf}, figurename=Figure, tablename=Table]{caption}

\usepackage[hidelinks]{hyperref}
\hypersetup{
  colorlinks   = true, 
  urlcolor     = esalightblue, 
  linkcolor    = esablue, 
  citecolor   = esablue 
}
\usepackage{amsfonts}       
\usepackage{amsmath}
\usepackage{amssymb}
\usepackage{mathtools}

\DeclarePairedDelimiter\floor{\lfloor}{\rfloor}
\usepackage[utf8]{inputenc} 
\usepackage[T1]{fontenc}    
\usepackage{hyperref}       
\usepackage{url}            
\usepackage{booktabs}       
\usepackage{nicefrac}       
\usepackage{lipsum}
\usepackage{graphicx}
\graphicspath{ {./images/} }
\usepackage[nameinlink, capitalise]{cleveref}
\usepackage{xcolor}
\definecolor{esablue}{RGB}{0,57,158}
\definecolor{esalightblue}{RGB}{0,155,219}
\definecolor{esared}{RGB}{150,1,54}
\usepackage{float}
\usepackage{algpseudocode}

\newtheorem{theorem}{Theorem}
\newtheorem{corollary}{Corollary}

\newtheorem{lemma}{Lemma}

\usepackage{todonotes}

\usepackage[giveninits=true,sorting=none,citestyle=numeric-comp]{biblatex}

%% file: math_abbrev.tex
\newcommand{\maxspikes}{N_\text{s}}
\newcommand{\indim}{d_\text{in}}
\newcommand{\encdim}{d_\text{enc}}
\newcommand{\decdim}{d_\text{dec}}
\newcommand{\outdim}{d_\text{out}}
\newcommand{\real}[1]{\mathbb{R}^{#1}}

\newcommand{\lp}[1]{L^{p}\left(#1\right)}
\newcommand{\thresh}{\vartheta}
\newcommand{\win}{\pmb{W}_\text{in}}
\newcommand{\wout}{\pmb{W}_\text{out}}
\newcommand{\spikein}{s_\text{in}}
\newcommand{\spikeout}{s_\text{out}}
\newcommand{\spikeset}[1]{S_{#1}}

%% file: content/abstract.tex
In a spiking neural network, is it enough for each neuron to spike at most once?
In recent work, approximation bounds for spiking neural networks have been derived, quantifying how well they can fit target functions.
However, these results are only valid for neurons that spike at most once, which is commonly thought to be a strong limitation.
Here, we show that the opposite is true for a large class of spiking neuron models, including the commonly used leaky integrate-and-fire model with subtractive reset: for every approximation bound that is valid for a set of multi-spike neural networks, there is an equivalent set of single-spike neural networks with only linearly more (or less) neurons, in the maximum number of spikes, for which the bound holds.
The same is true for the reverse direction too, showing that regarding their approximation capabilities in general machine learning tasks, single-spike and multi-spike neural networks are equivalent.
Consequently, many approximation results in the literature for single-spike neural networks also hold for the multi-spike case.

%% file: content/introduction.tex
Is it sufficient for spiking neurons to spike at most once, or is there any benefit of spiking multiple times?
How to best encode information in the temporal domain of spikes is still an open question. The possibilities range from encoding via exact spike times, such as in the latency of a neuron's spike response given a stimuli onset \cite{thorpe1996speed,gollisch2008rapid}, to averaged quantities such as spike time rates \cite{gerstner2014neuronal}.
Although using only single spikes might seem restrictive at first, it comes with several benefits.
First, for exact spike times, analytical gradient-based learning rules are available for specific neuron models \cite{kheradpisheh2022bs4nn,mostafa2017supervised,comsa2020temporal,goltz2021fast,klos2025smooth,stanojevic2023exact}, as well as learning of synaptic delays \cite{goltz2024delgrad} in addition to synaptic weights.
Second, decoding outputs using lowest latency leads to a low time-to-solution, while the inherent sparsity of these spiking neural networks (SNNs) promises high energy efficiency \cite{goltz2021fast,stanojevic2024high}.
This makes single-spike SNNs quite attractive for edge applications, as recently proposed in the context of artificial intelligence (AI) onboard spacecraft \cite{lunghi2025energy,izzo2023neuromorphic}.

For most practical purposes though, SNNs consisting of leaky integrate-and-fire (LIF) neurons (or variations thereof) with an unconstrained number of spikes per neuron are used nowadays.
These SNNs are then commonly trained using the surrogate gradient method \cite{neftci2019surrogate,eshraghian2023training,pehle2021norse}.
In fact, a common critique is that neurons which spike at most once restrict the expressivity of SNNs \cite{taylor2022robust,zenke2021visualizing,eshraghian2023training} and do not reflect biology well. 
This raises the question whether theoretical results derived for single-spike neurons apply to the more significant and relevant case of multi-spike neurons.

\begin{figure*}[h!]
    \centering
    \includegraphics[width=0.95\columnwidth]{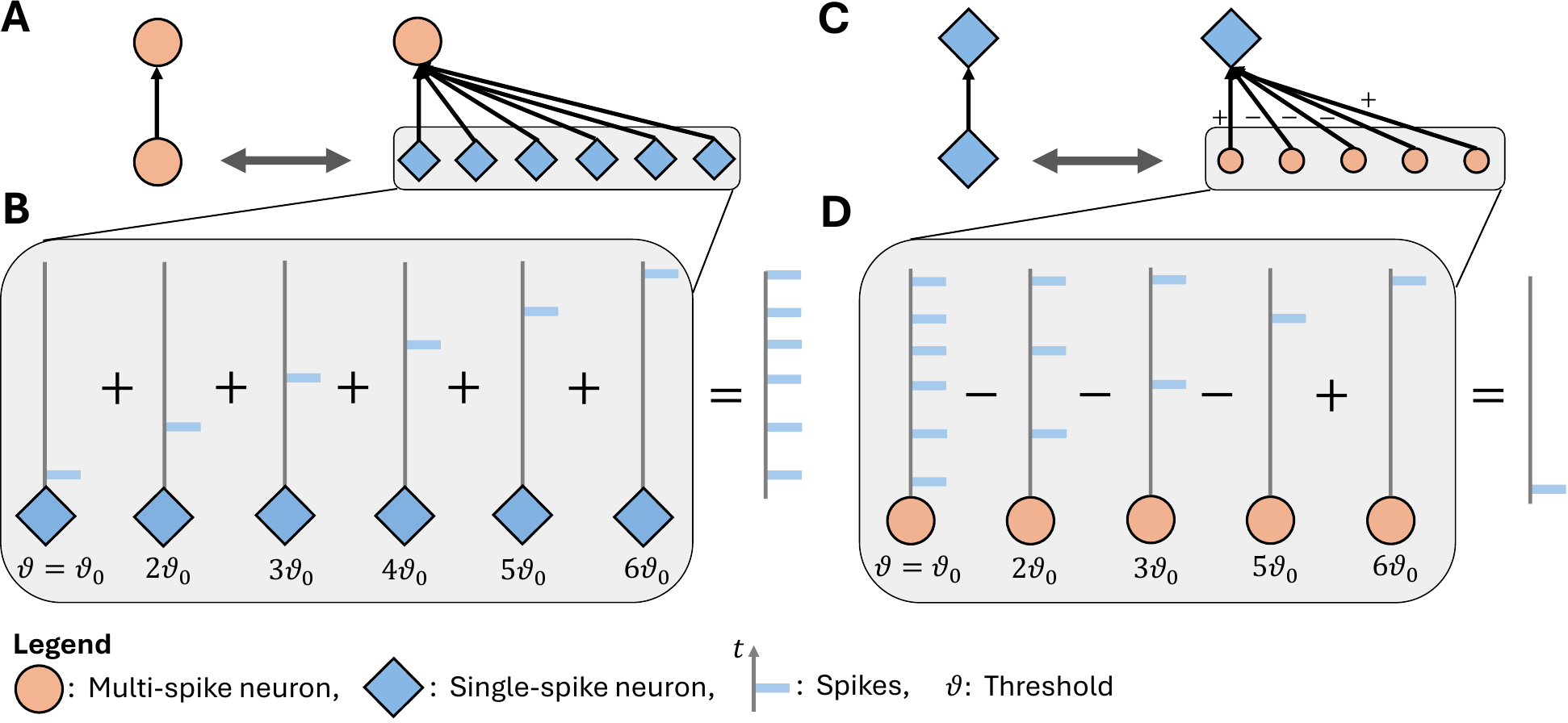}
	\caption{Illustration of how to construct equivalent single-and multi-spike SNNs, shown here for multi-spike neurons that spike at most six times in a given finite time interval. \textbf{A.} A multi-spike neuron can be replaced by a population of single-spike neurons that produce an equivalent spike output. \textbf{B.} The $k^\text{th}$ single-spike neuron spikes at the $k^\text{th}$ spike time of the multi-spike neuron. This is achieved by setting the threshold of each neuron accordingly (alternatively, the reset potential could be adjusted). \textbf{C.} A single-spike neuron can also be replaced by a population of multi-spike neurons. \textbf{D.} Given any input, the first neuron produces a spike train. We add additional neurons to the population that gain identical input, but have increasing thresholds. For instance, the second neuron only spikes at every second spike time of the first neuron. By choosing synaptic weights accordingly, all but the first spike of the first neuron are cancelled out, producing an effective spike train with only one spike.}
	\label{fig:illustration}
\end{figure*}

This work clarifies this open question, and we show a simple but extremely consequential result: both single-spike and multi-spike SNNs are equivalent when it comes to their approximation capabilities. 
Our result is inspired by recent work on the equivalence in approximation of fully-connected and convolutional neural networks (NNs) \cite{petersen2020equivalence}.
A formal statement of the result is given in \cref{theorem:main}, which is roughly summarised as follows:
\begin{enumerate}
    \item All upper bounds concerning approximation rates of multi-spike SNNs with $n \in \mathbb{N}$ neurons that spike at most $\maxspikes{} \in \mathbb{N}$ times translate to the same bounds concerning approximation rates of single-spike SNNs with at most $\maxspikes{} \cdot n$ neurons.
    \item All upper bounds concerning approximation rates of single-spike SNNs translate to the same bounds concerning approximation rates of multi-spike SNNs with at most $\alpha\maxspikes{} \cdot n$ neurons that spike at most $\maxspikes{}$ times.
    The prefactor satisfies $\alpha \leq \text{min}\left(1, \frac{6}{\pi^2} + 1/\sqrt{\maxspikes{}}\right)$ for $\maxspikes{} \geq 1$ and $\alpha < \frac{6}{\pi^2} + 1/(2\sqrt{\maxspikes{}})$ for $\maxspikes{} \geq 8$.
\end{enumerate}
A similar result holds for lower bounds (\cref{corrolary:lower}).
Both the single-spike and multi-spike NNs have to be based on the same neuron model for this transference principle to hold, i.e., the single-spike neurons behave dynamically like multi-spike neurons that stop spiking after the first spike.
The first part can be seen with the following direct argument: first, we replace each multi-spike neuron by $\maxspikes{}$ single-spike neurons, one per spike. Second, to ensure that the single-spike neurons spike at the correct time, their threshold is adjusted accordingly (\cref{fig:illustration}A,B), and we are done.
The central idea for showing the second statement is to combine the output of a small population of multi-spike neurons in such a way that all except one spike cancel out, allowing the population to mimic single-spike neurons (\cref{fig:illustration}C,D).
This result holds for a large family of spiking neuron models, including the widely used LIF neuron model with and without synaptic delays.

The result has far-reaching consequences, as every achievable approximation rate derived for single-spike SNNs automatically applies to multi-spike SNNs, and vice versa.
Thus, our result directly strengthens recently reported approximation bounds for SNNs, such as covering-number based upper learning bounds derived for a simple spike-response model \cite{neuman2024stable}, and lower bounds of the approximation error for non-leaky integrate-and-fire neurons (nLIF) derived using a piecewise decomposition of SNNs \cite{dold2025causal}.
Moreover, in \cite{stanojevic2023exact,stanojevic2024high}, it was shown that for a simple spike-response model with linearly rising postsynaptic potential, an exact mapping between ReLU NNs and SNNs exists, i.e., where both NNs produce the exact same output for every input sample.
Applying our result to this work, it follows that for each ReLU NN, there exists a multi-spike SNN (with the aforementioned neuron model) that satisfies the same approximation bounds.
Generalisation error bounds for ReLU NNs are well known and can, for instance, be derived using covering numbers or Rademacher complexity, while interpolation error bounds can be derived via the number of affine pieces \cite{petersen2024mathematical}.

In the following, we first provide an overview of all assumptions for deriving \cref{theorem:main}. We then continue to stating our main theorem, proving it, and briefly discussing the main limitations of our result for completeness.

%% file: content/background.tex
\subsection{Learning task} 

First, we introduce the set of all spike trains consisting of at most $N \in \mathbb{N}$ spikes: 
\begin{equation}
    \spikeset{N} \coloneqq \bigcup_{k=1}^{N} \left\{(t_1, \dots, t_k) \in [0, T]^k : t_1 \leq \dots \leq t_k\right\} \cup \{()\} \ \  \text{with} \ \  T \in \mathbb{R} \cup \{\infty\} \,,
\end{equation}
where $()$ is the empty sequence representing no spiking. In particular, we define $\spikeset{\infty} = \bigcup_{n \in \mathbb{N}} \spikeset{n}$ as the set containing all spike trains of finite length. From this, we define the set of all functions $f: \spikeset{\infty}^{\indim} \to \spikeset{\maxspikes}^{\outdim}$ mapping a set of $\indim \in \mathbb{N}$ input spike trains to $\outdim \in \mathbb{N}$ output spike trains with maximally $\maxspikes \in \mathbb{N}$ spikes per neuron. To ensure that SNNs can learn these functions, we enforce that $f$ is causal in time, i.e., $\text{min}\big(f(X)\big) > \text{min}(X)$ for all $X \in \spikeset{\infty}^{\indim}$.

To cover most standard learning scenarios encountered in the SNN literature, we compose these spike-train functions with a spike-encoder and decoder.
A spike-encoder is a function that translates the features of our data to spike trains, which then become the input to our SNN. Formally, we define encoders here as functions $E: \mathbb{R}^{\encdim} \to \spikeset{\infty}^{\indim}$, where $\encdim$ is the feature dimension of the data samples (e.g., the number of pixels of an image).
This general formulation covers established encoders, such as rate-and latency encoding \cite{eshraghian2023training,goltz2021fast}. In case of rate-encoding, the input features are real, positive values representing spike rates, which are translated to spike trains that feature these rates.
Similarly, for latency encoding, the inputs are the times-to-first-spike, which is translated into single-spike spike trains with this latency compared to a reference time.

Decoders do the opposite: take the output of the spike-based function and map it into a measurable space $\mathcal{O}$, $D: \spikeset{\maxspikes}^{\outdim} \to \mathcal{O}$, e.g., $\mathcal{O} = \real{\decdim}$ with $\decdim \in \mathbb{N}$.
Common decoders either work with exact spike times (latency, relative spike times, exact spike time matching) or derived quantities such as the membrane potential (matching membrane trace, mean membrane potential, maximum membrane potential) or spike counts \cite{eshraghian2023training}.

In summary, our results cover the case of general spike-based functions $f$  extended using common encoders and decoders: $f \mapsto D \circ f \circ E$.

\subsection{Spiking neuron models} 

This work applies at least to the family of spiking neuron models that satisfy the following three conditions:
\begin{enumerate}
    \item The neuron is threshold-based, i.e., spikes are generated when the neuron's membrane potential $u$ crosses a constant threshold value $\thresh$: $\{t: u(t) = \thresh\}$.
    \item After spiking, the neuron's membrane potential dynamics are unaffected apart from a constant offset $-\thresh$.
    \item Input currents are independent of a neuron's membrane potential, as for instance in case of current-based models (but not conductance-based ones when close to the reversal potentials).
\end{enumerate}
There is a large set of models satisfying these conditions, including the widely used leaky integrate-and-fire (LIF) neuron model without refractory period, variations thereof, such as the nLIF model \cite{mostafa2017supervised,goltz2021fast}, and other models such as the simple spike response model \cite{stanojevic2023exact,neuman2024stable}. 
For these models, the reset mechanisms that are commonly used also satisfy the second condition.
For example, especially for machine learning applications, the following type of LIF model is often used \cite{eshraghian2023training,pehle2021norse}:
\begin{equation}
    u(t+\mathrm{d}t) = - \beta u(t) + I(t) - s(t)\thresh \,,
\end{equation}
where $u(t)$ is the membrane potential of a single neuron at time $t$, $\beta > 0$ a constant, $I(t)$ is the time-dependent input to the neuron (e.g., currents caused by pre-synaptic spikes of other neurons), and $s(t) = \sum_i \delta(t-t_i)$ is a sum of Dirac delta functions representing the output spike times $t_i$ of the neuron.
Thus, whenever the neuron spikes, its immediate membrane potential is reduced by $\thresh$, as required by our second condition.
Although not specifically included here in our definition, the presented results apply to spiking neuron models with synaptic delays.

Models that do not necessarily satisfy these conditions are those with refractory periods, adaptive thresholds \cite{brette2005adaptive}, and spike generation without thresholds \cite{hodgkin1952quantitative,fitzhugh1961impulses,nagumo2007active}. This does not necessarily mean that the presented theorem does not hold for these neuron models, but merely that the presented proof cannot be applied in these cases.

\subsection{Network architectures}

The results in this work are stated for layered networks of single-and multi-spike neurons, allowing for connections between neurons of the same layer or neurons of different layers. We rule out autapses, i.e., recurrent connections of a neuron with itself. 
However, we further assume that network weights are bounded by $B$ in absolute value, with $B > 0$.
When talking about NNs in our main result, we always refer to this architecture.

\subsection{Bounded number of spikes}

Generally, we assume that spiking is constrained to a finite time interval $T$, leading to a maximum number $\maxspikes$ of times any multi-spike neuron can spike.
This does match realistic neurons, which also can only fire a finite number of times in a given time interval.
Moreover, for most spiking neuron models, with finite weights and non-zero membrane/synaptic time constants, the time-to-spike is finite, leading to an upper bound for the spike rate.

%% file: content/results.tex
Below, we state our main result: the equivalence of approximation of single-spike and multi-spike NNs (\cref{theorem:main}).
It holds for any spiking neuron models with simple reset dynamics, various spike time decoders, and synaptic delays (see \cref{section:background}).

\begin{theorem}[Transference principle] \label{theorem:main}
Let $\indim,\outdim,\maxspikes \in \mathbb{N}$, $f: \spikeset{\infty}^{\indim} \to \spikeset{\maxspikes}^{\outdim}$ be time-causal, and $F = D \circ f$ with decoder $D$. Furthermore, set $\Omega \subseteq  \spikeset{\infty}^{\indim}$. Then, the following hold for each $p \in (0, \infty]$ and each $\epsilon \geq 0$:

\begin{enumerate}
    \item If there is a multi-spike NN $\Psi$ with $n \in \mathbb{N}$ neurons and $L$ layers satisfying $\| F - D \circ \Psi \|_{\lp{\Omega}} \leq \epsilon$, then there is a decoder $D'$ and a single-spike NN $\Phi$ with $L$ layers and at most $\maxspikes\cdot n$ neurons, such that  $\| F - D' \circ \Phi \|_{\lp{\Omega}} \leq \epsilon$.
    \item If there is a single-spike NN $\Phi$ with $n \in \mathbb{N}$ neurons and $L$ layers satisfying $\| F - D \circ \Phi \|_{\lp{\Omega}} \leq \epsilon$, then there is a multi-spike NN $\Psi$ with $L$ layers and at most $\alpha \maxspikes\cdot n$ neurons, $\alpha \leq \text{min}\left(1, \frac{6}{\pi^2} + 1/\sqrt{\maxspikes{}}\right)$ for $\maxspikes{} \geq 1$ and $\alpha < \frac{6}{\pi^2} + 1/(2\cdot\sqrt{\maxspikes{}})$ for $\maxspikes{} \geq 8$, such that  $\| F - D \circ \Psi \|_{\lp{\Omega}} \leq \epsilon$.
\end{enumerate}

\end{theorem}

In other words: given a multi-spike NN, we can always find an equivalent single-spike NN that performs similarly on a given learning task and vice versa -- assuming that the dynamics of single and multi-spike neurons is the same (e.g., single-spike and multi-spike LIF with identical dynamics).
This result is obtained without changing the neuron models or the underlying network architecture.
Remarkably, for both directions, the number of neurons required to construct an equivalent NN increases, at most, linearly with the maximum number of spikes the multi-spike neurons can emit.

The following two extensions follow immediately from \cref{theorem:main}:

\begin{corollary}[Lower bounds]\label{corrolary:lower} \cref{theorem:main} extends also to lower bounds of approximation rates. Let $n \in \mathbb{N}$, $D$ a decoder, and $\maxspikes$, $\alpha$ as in Theorem 1. Then:
\begin{enumerate}
    \item If for all multi-spike NNs $\Psi$ with at most $\alpha \maxspikes \cdot n$ neurons and $L$ layers it holds that $\| F - D \circ \Psi \|_{\lp{\Omega}} \geq \epsilon$, then for all single-spike NNs $\Phi$ with $L$ layers and at most $n$ neurons, it holds that  $\| F - D \circ \Phi \|_{\lp{\Omega}} \geq \epsilon$.
    \item If for all single-spike NNs $\Phi$ with at most $\maxspikes\cdot n$ neurons, a decoder $D'$ as given in \cref{eq:dprime}, and $L$ layers it holds that $\| F - D' \circ \Phi \|_{\lp{\Omega}} \geq \epsilon$, then for all multi-spike NNs $\Psi$ with $L$ layers and at most $n$ neurons, it holds that  $\| F - D \circ \Psi \|_{\lp{\Omega}} \geq \epsilon$.
\end{enumerate}
\end{corollary}

\begin{corollary}[Encoders] \label{corollary:encoder}
    \cref{theorem:main,corrolary:lower} extend to common input spike-encoders as described in \cref{section:background}.
\end{corollary}

Thus, through our result, generalization and approximation bounds obtained for single-spike neuron models can be immediately mapped to the multi-spike case, such as bounds obtained using covering numbers \cite{neuman2024stable} and causal pieces \cite{dold2025causal}.
As we phrased \cref{theorem:main} for the rather general learning task of mapping from spike trains to spike trains, it applies both to static (e.g., image recognition) and temporal learning tasks (e.g., temporal sequence prediction) often studied with SNNs \cite{cramer2020heidelberg}.


\subsection{Proofs}

To prove \cref{theorem:main}, we first show how to replace a single neuron in a network, from which the theorem follows immediately by replacing all neurons in a network accordingly. 

\paragraph{Proof of the first part.} The first part of \cref{theorem:main} is straightforward. Assume a single multi-spike neuron with input weights $\win \in \real{d_0}$, output weights $\wout \in \real{d_1}$, and input spike trains $\spikein \in \spikeset{\infty}^{d_0}$.
Given the inputs, the neuron produces an output spike train $\spikeout \in \spikeset{\maxspikes}$ with at most $\maxspikes$ output spikes.
We can replace this neuron by at most $\maxspikes$ single-spike neurons with the same membrane dynamics, with neuron $k$ having threshold $k \cdot \thresh$ for $k \in [1, \maxspikes+1]$, and otherwise all neurons sharing the same input weights $\win$, output weights $\wout$, neuron parameters (e.g., time constants), and inputs $\spikein$ as the multi-spike neuron.
By \textit{the same membrane dynamics}, we mean that the single-spike neurons behave identical to the multi-spike neurons, just that they stop spiking after the first spike.

Assume now that $\spikeout = (t_1, \dots, t_N)$ for $N \leq \maxspikes$. The first single-spike neuron has the lowest threshold, and hence will spike at time $t_1$ and remain silent afterwards. 
At $t_1$, the multi-spike neuron is reset and its membrane potential will rise by $\thresh$ again until the second output spike time $t_2$.
Thus, at time $t_2$, the second single-spike neuron's membrane potential reaches its threshold $2\cdot \thresh$ and thus spikes.
Similarly, the third single-spike neuron spikes at $t_3$, the fourth at $t_4$, and so on.
Single-spike neurons $k > N$ are silent, as the remaining input spikes that occurred after $t_N$ do not drive the multi-and single-spike neurons across their thresholds.

Using this construction, any multi-spike neuron in $\Psi$ can be replaced by $\maxspikes$ single-spike neurons, and the resulting NN $\Phi$ will produce the exact same output spike times as $\Psi$ given any input spike trains.
However, the output format is now different, since spike times are scattered across neurons.
Assume that the $[(k-1) \cdot \maxspikes, k \cdot \maxspikes]$ single-spike neurons in the output layer of $\Phi$ represent the $k^\text{th}$ multi-spike neuron in the last layer of $\Psi$.
Denote by $s_i \in \spikeset{1}$ the output of the $i^\text{th}$ neuron in the output layer of $\Phi$.
By choosing the decoder $D' = D \circ A$ with 
\begin{equation}\label{eq:dprime}
    A(X)_k = s_{(k-1)\maxspikes} \oplus \dots \oplus s_{k \cdot \maxspikes} \,,
\end{equation}
for $X \in \spikeset{1}^{\outdim \cdot \maxspikes}$, we then arrive at the second statement of \cref{theorem:main}.
Here, we introduce a notation for appending sequences:  $(t_0, \dots, t_K) \oplus (t_{K+1},\dots,t_N) = (t_0, \dots, t_K, t_{K+1}, \dots, t_N)$, and $(t_0, \dots, t_K) \oplus () = (t_0, \dots, t_K)$.

In the above proof, we assumed no synaptic delays. Assume $\pmb \delta_\text{in} \in \mathbb{R}^{d_0}_+$ are input delays, meaning that the spike times of the $i^\text{th}$ input spike train are all shifted by $\delta_{\text{in},i}$. 
Further assume that $\pmb \delta_\text{out} \in \mathbb{R}^{d_1}_+$ are output delays, meaning that output spikes transmitted to the $j^\text{th}$ post-synaptic neuron are delayed by $\delta_{\text{out},j}$.
As long as the single-spike neurons have the same input and output delays as the multi-spike neuron (similarly to how they share the same input and output weights), the above proof applies.

\paragraph{Proof of the second part. } The second part of \cref{theorem:main} can be derived using a similar idea. Instead of using multiple neurons to create a spike train, we use them to ensure that, in the relevant time period, this population of multi-spike neurons emits at most one effective spike, i.e., only one spike that affects its postsynaptic partners.

We take the same setup as before, but start from a single-spike neuron that produces outputs $\spikeout \in \spikeset{1}$.
We reproduce its input-output mapping using a population of at most $\maxspikes$ multi-spike neurons that have the same membrane dynamics as the single-spike neuron, as well as the same inputs, input weights, and input and output delays.
In the following, without loss of generality and to simplify the notation, we drop the delays.
Then, we set the threshold of the $k^\text{th}$ multi-spike neuron to $k \cdot \thresh$, and its output weights to $\alpha_k \cdot \wout$.
All other neuron parameters are identical to the ones of the single-spike neuron.

Assume that given some input, the single-spike neuron spikes at time $t_1$. 
We choose as our first neuron in the multi-spike population one that spikes at times $s_1 = (t_1, \dots, t_{\maxspikes})$, where $t_2, \dots, t_{\maxspikes}$ are sorted and larger than $t_1$, but otherwise arbitrary.
Then the second neuron in the multi-spike population, with threshold $2\cdot \thresh$, spikes only every second time $s_2 = \left(t_2, t_4, \dots, t_{2\cdot \floor{\maxspikes/2}}\right)$, with $\floor{\cdot}$ denoting rounding down.
Similarly, the third neuron in the multi-spike population spikes only at times $s_3 = \left(t_3, t_6, \dots, t_{3\cdot \floor{\maxspikes/3}}\right)$.
Generally, since the first multi-spike neuron spikes $\maxspikes$ times in the time period $[0, T]$, the $k^\text{th}$ neuron spikes only $\floor{\maxspikes/k}$ times.
To have the same effect on the postsynaptic neurons as the single-spike neuron, we thus need:
\begin{equation}
    \sum_{i = 1}^{\maxspikes} \sum_{t' \in s_i} \alpha_i \wout \delta(t-t') = \wout \delta(t-t_1) \,,
\end{equation}
which is equivalent to the condition $\pmb M \pmb \alpha = \pmb e_0$ with
\begin{equation}
    M_{ij} = \begin{cases}
    1 & \text{if}\ i \geq j\ \text{and}\ (i+1)\%(j+1) = 0 \,, \\
    0 & \text{otherwise} \,,
    \end{cases}
\end{equation}
with $\%$ being the modulo operator, $\alpha = \left(\alpha_1, \dots, \alpha_{\maxspikes}\right)$, and $\pmb e_0 = (1, 0, \dots, 0)$ the unit vector with all zeroes except the first element.
For example, for $\maxspikes = 6$, the linear equation is:
\begin{equation}
    \begin{pmatrix}
1\ & 0\ & 0\ & 0\ & 0\ & 0 \\
1\ & 1\ & 0\ & 0\ & 0\ & 0 \\
1\ & 0\ & 1\ & 0\ & 0\ & 0 \\
1\ & 1\ & 0\ & 1\ & 0\ & 0 \\
1\ & 0\ & 0\ & 0\ & 1\ & 0 \\
1\ & 1\ & 1\ & 0\ & 0\ & 1
\end{pmatrix}
\begin{pmatrix}
\alpha_1 \\ \alpha_2 \\ \alpha_3 \\ \alpha_4 \\ \alpha_5 \\ \alpha_6
\end{pmatrix}
=
\begin{pmatrix}
1 \\ 0 \\ 0 \\ 0 \\ 0 \\ 0
\end{pmatrix}.
\end{equation}
In other words: we have to find appropriate weights $\alpha_i$ such that all spikes except the first one at time $t_1$ cancel each other.
From this, it follows immediately that $\alpha_1 = 1$.
Furthermore, since $\pmb M$ is a lower-triangular matrix with non-zero diagonal elements, this linear equation always has a unique solution.

This way, we can replace every neuron in the single-spike NN $\Phi$ by at most $\maxspikes$ multi-spike neurons and appropriately chosen weights to get an equivalent multi-spike NN $\Psi$.
For neurons in the output layer, we choose a decoder $D$ that filters out the first spike of each neuron to conclude the proof.
The solution for the weight factors is given by \cref{lemma:weights}.

\begin{lemma}(Weight factors)\label{lemma:weights}
Assume we want to replace a single-spike neuron with $\maxspikes$ multi-spike neurons.
Let $n\in [1, \maxspikes]$. Furthermore, let $n = \prod_{i=1}^{k} p_i^{m_i}$ be its prime factorization with $k \in \mathbb{N}$, $k > 0$, $m_i \in \mathbb{N}$, and $p_i \leq n$ being prime numbers.
Choose $\alpha_1 = 1$. If for every $n > 1$, we choose the weight factors
\begin{equation}
    \alpha_n = \begin{cases}
    0 & \text{if}\ \mathrm{max}\left(\{m_i : i \in [1,k]\}\right) > 1 \,, \\
    (-1)^k & \text{otherwise} \,,
    \end{cases}
\end{equation}
then the vector $\pmb \alpha = \left(\alpha_1, \dots, \alpha_{\maxspikes}\right)$ solves $\pmb M \pmb \alpha = \pmb e_0$.
\end{lemma}
\cref{lemma:weights} is summarised as follows: a single-spike neuron can be replaced by $\eta(\maxspikes{})$ multi-spike neurons that spike at most $\maxspikes$ times in a given time window $[0,T]$, where $\eta(\maxspikes{})$ is the number of square-free numbers smaller or equal than $\maxspikes$ (i.e., we remove all neurons $j$ with weight factor $\alpha_j = 0$).
A number is square-free if in its prime factorization, all factors are unique, i.e., do not appear multiple times.
$\eta(\maxspikes{})$ is upper bounded by $\eta(\maxspikes{}) < \zeta(2)^{-1} \cdot \maxspikes{} + \sqrt{\maxspikes{}}$ for $\maxspikes{} \geq 1$ and $\eta(\maxspikes{}) < \zeta(2)^{-1} \cdot \maxspikes{} + \frac{1}{2}\sqrt{\maxspikes{}}$ for $\maxspikes{} \geq 8$, with $\zeta(2)^{-1} = 6/\pi^2$ \cite{moser1966error}.
In addition, we always have $\eta(\maxspikes{}) \leq \maxspikes{}$. Combined, this results in the estimate for the required number of neurons given in \cref{theorem:main}.
Particularly, in the limit of large $\maxspikes{}$, we have $\lim_{\maxspikes{} \to \infty} \eta(\maxspikes{})/\maxspikes{} = 6/\pi^2$.

\begin{figure*}[t!]
    \centering
    \includegraphics[width=0.95\columnwidth]{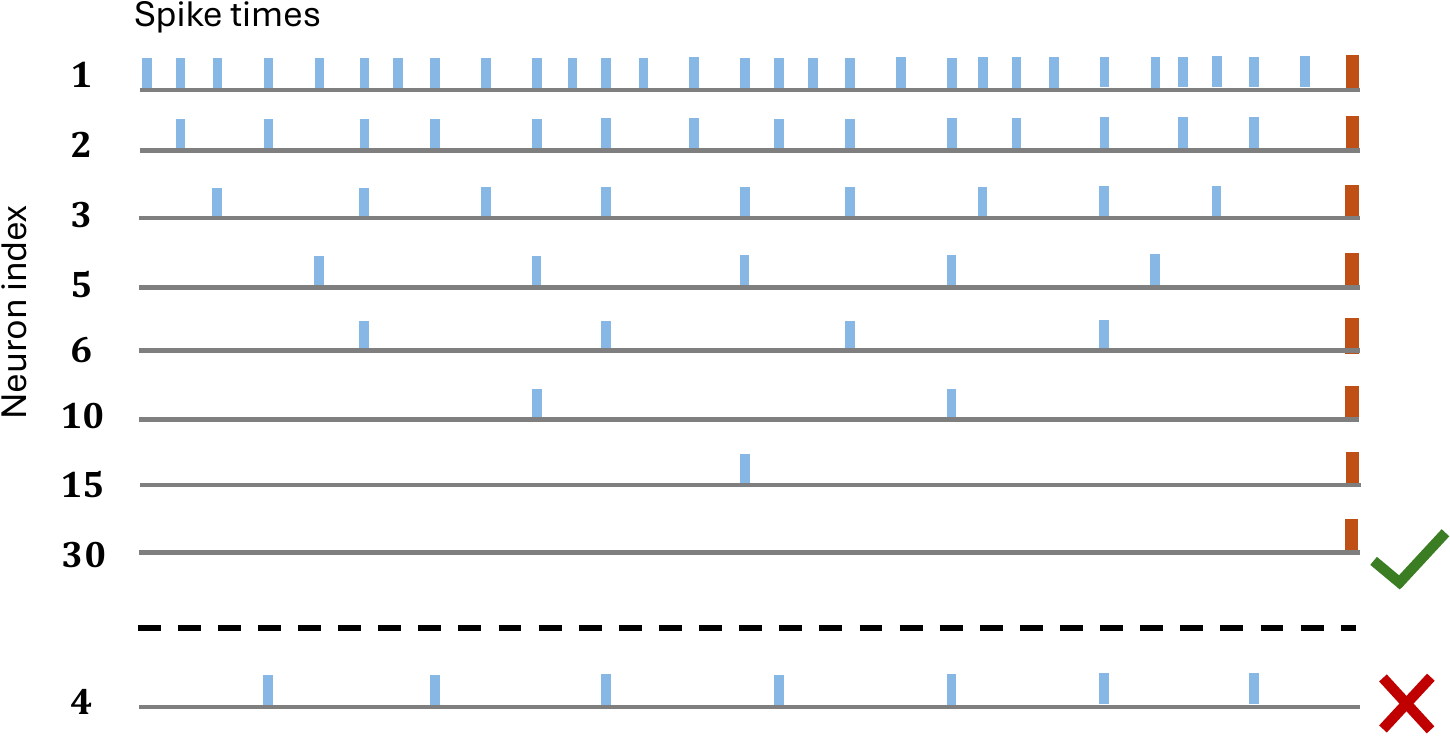}
	\caption{Illustration of how to effectively cancel out spike times. We assume a population of neurons where the neuron with index $i$ spikes at every $i^\text{th}$ spike time of the first neuron with index $1$. Given the $n^\text{th}$ spike time (here: $n = 30$), only those neurons whose index divides $n$ without remainder also spike at this time. Thus, we only have to consider the outgoing weights of this subset of neurons for cancelling the $n^\text{th}$ spike.}
	\label{fig:cancel}
\end{figure*}

\paragraph{Proof idea. } Before formalising the proof, we briefly illustrate the main idea.
Assume we have $\maxspikes$ multi-spike neurons.
The neuron with index $1$ spikes exactly $\maxspikes$ times.
The neuron with index $2$ spikes at every second spike time of the neuron with index $1$, the third at every third spike time of the neuron with index $1$, and so on (see \cref{fig:cancel}).
We now ask: which neurons with index $> 1$ spike at the $n^\text{th}$ time?
In the end, for these neurons we have to choose the outgoing weights in a way that the effect of all spikes cancels, meaning that this population of neurons produces \textit{effectively} no spike at the $n^\text{th}$ time.

For example, assume $n = 30$.
For a neuron to spike at the $30^\text{th}$ spike time, its index has to divide $n$ without remainder.
In our example, this means that the neurons with index $\mathcal{D} = \{2, 3, 5, 6, 10, 15, 30\}$ spike at this time -- all other neurons do \textbf{not} spike at the $30^\text{th}$ time (\cref{fig:cancel}).
As we are looking for dividing factors, we can simplify by looking at the prime factorisation of $n = 30 = 1 \cdot 2 \cdot 3 \cdot 5$. $\mathcal{D}$ is then obtained by forming the products over all possible subsets, i.e.,  $\mathcal{D} = \{2, 3, 5, 2\cdot 3, 2\cdot5, 3\cdot 5, 2\cdot 3 \cdot 5\}$.
The assigned weights for a neuron with index $m$ is given by $(-1)^{k_m}$, where $k_m$ is the number of prime factors $m$ composes to.
Thus, we just have to count:
we have $\binom{3}{1} = 3$ neurons whose indices are already a prime number (2,3,5), $\binom{3}{2} = 3$ neurons whose indices decompose into a product of two primes (6, 10, 15), and $\binom{3}{2} = 1$ neuron whose index decomposes into a product of three primes (30).
Summing up these spikes with their corresponding weights, we get $\binom{3}{1} (-1)^1 + \binom{3}{2} (-1)^2 + \binom{3}{3} (-1)^3 = -3 + 3 - 1 = -1$. Adding the spike of the neuron with index $1$ (and weight $\alpha_1 = 1$), all spikes cancel perfectly out ($-1 + 1 = 0$).

In this example, $n = 30$ was a square-free number, meaning that no prime number appears multiple times in its factorisation.
However, since all the neurons that were involved in cancelling the $30^\text{th}$ spike time will also spike at all multiples of $30$, they also cancel out at those times.
For example, assume $n =60 = 1 \cdot 2^2 \cdot 3 \cdot 5$.
We already showed that the neurons with indices $\{2, 3, 5, 6, 10, 15, 30\}$ cancel out the $30^\text{th}$ spike time of the neuron with index $1$. 
Hence, they also cancel out the $60^\text{th}$ spike time, meaning we do not need an additional neuron with index $60$ at this time (i.e., we either set $\alpha_{60} = 0$, or we simply remove this neuron from the population altogether).
This is true for all neurons whose index is not a square-free number.
Hence, the number of neurons required to cancel out $\maxspikes$ spikes is given by the number of square-free numbers smaller or equal to $\maxspikes$.
In this work, we cite two standard bounds for this, with the second one being tighter but also only valid for larger values of $\maxspikes$.

\paragraph{Proof. } Assume the $n^\text{th}$ spike time $t_n$, with $1 < n \leq \maxspikes$. 
Let $n = \prod_{i=1}^{k} p_i^{m_i}$ be its prime factorization.
First, we isolate all multiple factors by defining $\beta = \prod_{i=1}^{k} p_i^{m_i-1}$, resulting in $n = \beta \cdot \prod_{i=1}^{k} p_i$.
Note then that for any $C \subseteq \{1, \dots, k\}$ (with $\bar{C}$ being its complement), we have $n = L_C \cdot \beta \prod_{i \in \bar{C}} p_i$, with $L_C = \prod_{i \in C} p_i$, which means that the neuron with index $L_C$, which emits a spike at every $L_C^\text{th}$ spike time, also spikes at the $n^\text{th}$ spike time ($n$ is a multiple of its period $L_C$).
This is also true for neurons with indices whose prime factorization is not square-free, but according to \cref{lemma:weights}, their weight factors are set to 0 and we can neglect them.
As an example, assuming $n = 60$ again, one possible arrangement is $\beta = 2$ and $C = \{2,3\}$, leading to $L_C = 2\cdot 3 = 6$.

Thus, for the $n^\text{th}$ spike time, output spikes of all neurons with indices that decompose into products of subsets of the prime numbers $\{p_1, \dots, p_k \}$ are summed up.
There are, in total, $\sum_{i=1}^{k} \binom{k}{i}$ such combinations, and therefore $\sum_{i=1}^{k} \binom{k}{i}$ spikes that overlap.
Following \cref{lemma:weights}, the weight factor for each neuron's spike only depends on the number of primes its index factors in.
Thus, the weighted sum of all neurons is given by $\sum_{i=1}^{k} \binom{k}{i} (-1)^i = \sum_{i=0}^{k} \binom{k}{i} (-1)^i 1^{k-i} - 1 = (1-1)^k - 1 = -1$, which perfectly cancels the spike of the first neuron with $\alpha_1 = 1$.
Since this is valid for all values of $n$, the proof is complete.

\paragraph{Corollaries.} Both $\Psi$ and $\Phi$ produce the exact same spike times as output given the same spike trains as input, from which both corollaries follow immediately.

%% file: content/limitations.tex
We would like to stress that \cref{theorem:main} is not universally valid for all spiking neuron models.
Formally, we only prove it for models satisfying the requirements stated in \cref{section:background} -- although we do not rule out that the theorem holds true for other neuron models as well.

A major type of neurons not covered is those with a hard refractory period.
In this case, proving the direction multi-spike $\rightarrow$ single-spike in \cref{theorem:main} becomes very hard, if not impossible.
For example, for the proof, we require that every single-spike neuron has, up to a constant offset, the same membrane trace as the multi-spike neuron until the threshold is crossed.
However, during a hard refractory period, the multi-spike neuron's membrane potential is clamped to a constant value, while all single-spike neurons that haven't spiked yet continue to follow their inputs -- thus diverging from the multi-spike neuron and potentially leading to differences in the spike times.
The same applies for other effects that change a neuron's membrane dynamics after it spiked, such as adaptive thresholds.

Generally, \cref{theorem:main} can still be shown to be valid as long as the effect induced by spiking on the membrane dynamics can be modelled via an ordinary self-connection, such as certain soft refractory period mechanisms.

%% file: content/acknowledgments.tex
D.D. was funded by the Horizon Europe's Marie Skłodowska-Curie Actions (MSCA) Project 101103062 (BASE).